\pdfoutput=1

\documentclass[11pt]{article}
\usepackage{authblk}

\usepackage{EMNLP2022}

\usepackage{times}
\usepackage{latexsym}

\usepackage[T1]{fontenc}

\usepackage[utf8]{inputenc}

\usepackage{microtype}

\usepackage{inconsolata}

\usepackage{xspace}

\newcommand{\bart}{\textsc{Bart}\xspace}
\newcommand{\tldr}{\textsc{tl;dr}\xspace}
\newcommand{\rgo}{\textsc{rouge-1}\xspace}
\newcommand{\rgt}{\textsc{rouge-2}\xspace}
\newcommand{\rgl}{\textsc{rouge-l}\xspace}
\newcommand{\loss}{\ell}
\newcommand{\losslrg}{\mathcal{L}}

\newcommand{\currsum}{\textsc{CurrSum}}

\newcommand{\superl}{\textsc{SuperLoss}}

\newcommand\mentsum{\textsc{MentSum}\xspace}

\usepackage{bibentry}

\usepackage{times}
\usepackage{latexsym}

\usepackage{booktabs}

\usepackage{textcomp}
\usepackage{graphicx}
\usepackage{multirow} 
\usepackage{enumitem}

\usepackage{amsmath}
\usepackage{xcolor}

\usepackage{upquote}
\DeclareMathOperator*{\argmin}{arg\,min}
\usepackage{arydshln}

\usepackage{etoolbox}
\makeatletter
\patchcmd{\maketitle}
 {\def\@makefnmark}
 {\def\@makefnmark{}\def\useless@macro}
 {}{}
\makeatother

\title{Curriculum-guided Abstractive Summarization \\for Mental Health Online Posts}

\author[1*]{\textbf{Sajad Sotudeh} \thanks{* Work partially done during the internship at Adobe Research.}}
\author[1]{\textbf{Nazli Goharian}}

\author[2]{\textbf{Hanieh Deilamsalehy}}
\author[2]{\textbf{Franck Dernoncourt}}

\affil[1]{IRLab, Georgetown University}
\affil[ ]{{\fontfamily{lmtt}\selectfont{ \{sajad,nazli\}@ir.cs.georgetown.edu}}}

\affil[2]{Adobe Research} 
\affil[ ]{{\fontfamily{lmtt}\selectfont{
\{deilamsa,franck.dernoncourt\}@adobe.com}
}}

\begin{document}
\maketitle
\begin{abstract}

Automatically generating short summaries from users' online mental health posts could save counselors' reading time and reduce their fatigue so that they can provide timely responses to those seeking help for improving their mental state. Recent Transformers-based summarization models have presented a promising approach to abstractive summarization. They go beyond sentence selection and extractive strategies to deal with more complicated tasks such as novel word generation and sentence paraphrasing. Nonetheless, these models have a prominent shortcoming;  their training strategy is not quite efficient, which restricts the model's performance. In this paper, we include a curriculum learning approach to reweigh the training samples, bringing about an efficient learning procedure. We apply our model on \textit{extreme} summarization dataset of \mentsum posts ---a dataset of mental health related posts from Reddit social media. Compared to the state-of-the-art model, our proposed method makes substantial gains in terms of \textsc{Rouge} and \textsc{Bertscore} evaluation metrics, yielding 3.5\% (\rgo), 10.4\% (\rgt), and 4.7\% (\rgl), 1.5\% (\textsc{bertscore}) relative improvements.

\end{abstract}

\section{Introduction}

Summarization of mental health online posts is an emerging task that aims to summarize users' posts who are seeking help to enhance their mental state in online networks such as Reddit~\footnote{\url{https://www.reddit.com/}} and Reachout~\footnote{\url{https://au.reachout.com/}}. The post might address several issues of the user's concerns or simply be an elaboration on the user's mental and emotional situation. With user preference, each user-written post can be accompanied by a succinct summary (known as \tldr~\footnote{\tldr{} is the abbreviation of ``Too Long, Didn't Read''. We use ``\tldr{}'' and ``summary'' exchangeably in this paper.}), condensing major points of the user post. This \tldr{} summary is deemed to urge the counselors for a faster read of the user's posted content before reading the post in its entirety; hence, counsellors can provide responses promptly. Herein, we aim to improve state-of-the-art results reported in \cite*{Sotudeh2022MentSumAR} for this task.


Large-scale deep neural models are often hard to train, leaning on intricate heuristic set-ups, which can be time-consuming and expensive to tune~\cite{Gong2019EfficientTO, Chen2021EarlyBERTEB}. This is especially the case for the Transformers-based summarizers, which have been shown to consistently outperform the RNN networks when rigorously tuned~\cite{Popel2018TrainingTF}, but also require heuristics such as specialized learning rates and large-batch training~\cite{Platanios2019CompetencebasedCL}. In this paper, we attempt to overcome the mentioned problem on \bart{}~\cite{Lewis2020BARTDS} Transformers-based summarizer by introducing a \textit{Curriculum Learning (CL)} strategy~\cite{Bengio2009CurriculumL} for training the summarization model, leading to improved convergence time, and performance. 

Inspired by humans' teaching style, \textit{curriculum learning} suggests moving the teaching process from easier samples to more difficult ones and dates back to the nineties~\cite{Elman1993LearningAD}. The driving idea behind this approach is that networks can accomplish better task learning when the training instances are exposed to the network in a specific and certain order, from easier samples to more difficult ones~\cite{Chang2021DoesTO}. In the context of neural networks, this process can be thought of as a technique that makes the network robust to getting stuck at local optima, which is more likely in the early stages of the training process. Given the mentioned challenge of summarization networks, we utilize the \superl~\cite{Castells2020SuperLossAG} function that falls into the family of confidence-aware curriculum learning techniques, introducing a new parameter called confidence (i.e., $\sigma$) to the network. 
We validate our model on \mentsum~\cite*{Sotudeh2022MentSumAR} dataset,  containing over 24k instances mined from 43 mental health related communities on Reddit social media. Our experimental results show the efficacy of applying curriculum learning objectives on \bart{} summarizer, achieving a new state-of-the-art performance. 
\section{Related Work}
While majority of works in mental health research have focused on studying users' behavioral patterns through classification and prediction tasks~\cite{Choudhury2013PredictingDV,Resnik2013UsingTM,Coppersmith2014MeasuringPT,Yates2017DepressionAS,Cohan2017TriagingCS,Cohan2018SMHDAL,MacAvaney2018RSDDTimeTA}, summarization of online mental health posts has been recently made viable. ~\citet*{Sotudeh2022MentSumAR} were the first to step on this direction via introducing \mentsum dataset of online mental health posts, pinpointing the baseline results.  Curriculum Learning (CL) has gained growing interest from the research communities during the last decade~\cite{Tay2019SimpleAE, MacAvaney2020TrainingCF, Xu2020CurriculumLF}.
\citet{Bengio2015ScheduledSF} were the first to apply this strategy in the context of sequence prediction through \textit{scheduled sampling} approach, which gently changes the training process from ground truth tokens to model-generated ones during decoding. 
Sample's \textit{difficulty} is a key concept in this scheme as it is used to distinguish easy examples from difficult ones. Researchers have used many textual features as the ``difficulty measure'' including n-gram frequency~\cite{Haffari2009ActiveLF}, word rarity and sentence length~\cite{Platanios2019CompetencebasedCL}. Recent works~\cite{Saxena2019DataPA, Cachola2020TLDRES} have made use of confidence-aware approaches that learn the difficulty of training samples and dynamically reweight samples in the training process.

\section{Our Approach}
In this section, we describe the details of our proposed model, where a curriculum learning architecture is added to the \bart 's Transformers-based framework, upweighting easier training samples; hence, increasing their contribution in learning stage. 

\begin{figure}
    \centering
     \includegraphics[scale=0.3]{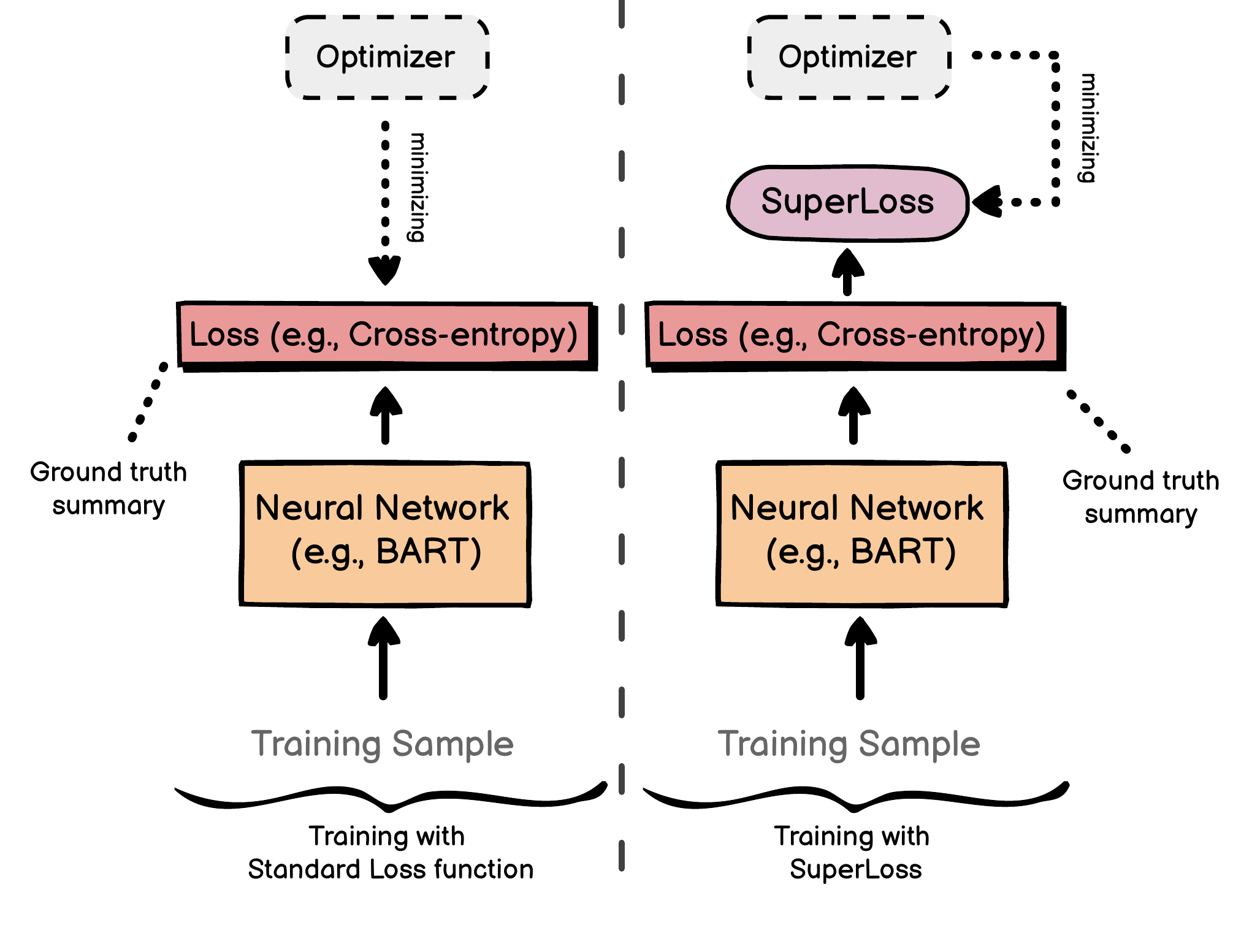}
    \caption{Training with standard loss function (left-hand side) and SuperLoss criteria (right-hand side)}. 
    \label{fig:superloss}
    \vspace{-.6em}
\end{figure}

\noindent \textbf{Curricular Learner for \bart{}.}
Considering the applicability of curriculum learning in training large-scale networks, we aim to use it in our summarization framework. Before incorporating the curriculum learning strategy into our model's training stage, we first need to define the \textit{difficulty} metric to distinguish the hardness of samples. In practice, estimating a prior difficulty for each sample is considered a complex task, so we propose to discriminate the samples with progressive signals ---such as the respective sample loss at each training iteration--- in the training process. In this context, CL is achieved by predicting the difficulty of each sample at the training iterations in the form of a weight, such that difficult samples receive lower weights during the early stages of training and vice versa. To model the curriculum, we propose to use \superl~\cite{Castells2020SuperLossAG} which is a generic loss criterion built upon the task loss function 
as shown in Figure~\ref{fig:superloss}.

More specifically, \superl{} is a task-agnostic confidence-aware loss function that takes in two parameters: (1) the task loss $\losslrg_i =  \loss(y_i, \widehat{y_i})$, where $y_i$ is neural network's (i.e., \bart's generated summary) output and $\widehat{y_i}$ is the gold label (i.e., ground-truth summary); and (2) $\sigma_i$ as the confidence parameter of the $i$th sample. \superl{} is framed as $\text{L}_{\lambda}(\losslrg_i, \sigma_i)$ and computed as follows,

\begin{equation}
    \text{L}_{\lambda}(\losslrg_i, \sigma_i)= (\losslrg_i - 	\tau) \sigma_i + \lambda (\log \sigma_i)^2
\end{equation}
in which $\lambda$ is the regularization parameter, and $\tau$ is the running or static average of task loss (i.e., $\losslrg$) during the training. While \superl{} provides a well-defined approach to curriculum learning strategy, learning $\sigma$ parameter is not tractable for tasks with abundant training instances such as text summarization. To circumvent this issue and hinder imposing new learnable parameters, \superl{} suggests using the converged value of $\sigma_i$ at the limit,

\begin{align}
\label{eqn:eqlabel}
\begin{split}
\sigma^{*}_{\lambda} (\loss_i) &= \argmin_{\sigma_i} \text{L}_{\lambda} (\loss_i, \sigma_i)
\\
 \text{SL}_\lambda(\loss_i) &= \text{L}_\lambda(\loss_i, \sigma^{*}_{\lambda}(\loss_i, \sigma_i)) = \min_{\sigma_i} \text{L}_\lambda (\loss_i, \sigma_i),
\end{split}
\end{align}

Using this technique, the confidence parameters are not required to be learned during the training. \citet{Castells2020SuperLossAG} found out that $\sigma^{*}_{\lambda} (\loss_i)$ has a closed-form solution, computed as follows,

\begin{equation}
     \sigma^{*}_{\lambda} (\loss_i) = e ^ {-W (\frac{1}{2} \max (-\frac{2}{e}, \beta))}, \beta = \frac{\loss_i - \tau}{\lambda}
\end{equation}
in which $W$ is the Lambert W function. With this in mind, \superl{} upweights easier samples dynamically during the training, providing a curriculum learning approach to summarization. We call this model \currsum{} in our experiments. 

\section{Experimental Setup}
\label{sec:res}

\subsection{Dataset}
We use the \mentsum dataset in our experiments. This dataset contains over 24k post-\tldr{} pairs, making up 21,695 (train), 1209 (validation), and 1215 (test) instances, and is gathered by mining 43 mental health subreddit communities on Reddit with rigorous filtering rules. We refer the readers to the main paper for more details on this dataset~\cite*{Sotudeh2022MentSumAR}. 

\subsection{Comparison}
We compare our model against the \bart{}~\cite{Lewis2020BARTDS} baseline, which does not utilize the curriculum learning objective. \bart is an abstractive model that uses a pre-trained encoder-decoder architecture, unlike \textsc{Bert} which only utilizes a pre-trained encoder. As shown in \cite*{Sotudeh2022MentSumAR}, \bart{} is the strongest baseline; thus, we apply CL on it to evaluate its impact on summarization. We refer the reader to the original paper for more extractive and abstractive baselines.

\subsection{Implementation details}
We use the Huggingface's Transformers library ~\cite{Wolf2020Transformers}~\footnote{\url{https://github.com/huggingface/transformers}} to implement our models.
We train all of our models for 8 epochs, performing evaluation step in intervals of 0.5 epochs, and use the checkpoint that achieves the best \textsc{Rouge-L} score in the validation for the inference. 
AdamW optimizer~\cite{Loshchilov2019DecoupledWD} initialized with learning rate of $3e-5$, $(\beta_1, \beta_2)= (0.9, 0.98)$, and weight decay of 0.01 is used for all of our summarization models, as well as for \textsc{Bart}. 
Cross-entropy loss is used for all models. 
To keep track of the learning process, we use Weights \& Biases~\cite{wandb} toolkit~\footnote{\url{https://wandb.ai/}}. 

\section{Ethics and privacy}
While we use publicly available Reddit data to train and validate our model, we recognize that special care should be taken when dealing with this type of data due to its sensitivity and users' privacy~\cite{Benton2017EthicalRP,Nicholas2020EthicsAP}. Thus, we have made no attempts to identify, contact the users, or explore user relations to trace their social media account. It has to be mentioned that the \mentsum{} dataset is distributed through Data Usage Agreement (DUA)~\footnote{\url{https://ir.cs.georgetown.edu/resources/mentsum.html}} document to further ensure that users' identity would never be disclosed.

\begin{table}[t]
\centering
\scalebox{0.8}{
 
   \begin{tabular}{l@{\hspace{1.1cm}}rrrr}
       \toprule
           
     Model                    & {R-1}  & {R-2}  & {R-L} & {BS} \\
    \midrule
       \textsc{OracleExt}             &  35.98 & 11.59 & 23.21 & 82.72 \\



     \textsc{Bart}~\citeyearpar{Lewis2020BARTDS} & 29.13 & 7.98 & 20.27 & 85.01  \\
     
    \hdashline[2pt/2pt] \vspace{-0.8em} \\
    \currsum{} (Ours) & \textbf{30.16}& \textbf{8.82}& \textbf{21.24} & \textbf{86.32} \\
    \bottomrule  
\end{tabular}

}

\caption{\textsc{Rouge} and \textsc{Bertscore} results on test set of \mentsum dataset. As \bart{} was the most performant baseline provided in ~\cite*{Sotudeh2022MentSumAR}, we evaluate the effectiveness of Curriculum on \bart in this work. For other baselines, we refer the reader to the main paper.
}
\label{tab:final}
\vspace{-1em}

\end{table}
\section{Results}
\begin{table*}[t]
\centering\small
\scalebox{.8}{
\begin{tabular}{ p{0.3\linewidth}  p{0.4\linewidth}  p{0.4\linewidth}}
\toprule
       Human-written & \bart-generated & \currsum-generated \\ \midrule
      
      
     \vspace{.25em}
      \textit{... having kids ... my ADHD ...  opinions on parenthood...} 

      & 
          \vspace{.25em}

      \footnotesize \textit{I think it's better to \textcolor{gray}{not have a child} now than \textcolor{gray}{never having a child} in the future.}
      
      &
          \vspace{.25em}

      \textit{I'm \textbf{thinking of having a child} and I \textbf{ don't want to be a shi**y dad} for my \textbf{ADHD}. Any \textbf{parents with ADHD experience?}}
          \vspace{.25em}

      \\
      
      \midrule
      
      
          \vspace{.25em}

      \textit{... symptoms of depression ... go to get help ...}
      
      &
          \vspace{.25em}

      \textit{I think I \textbf{am clinically depressed}, and I have pretty much \textbf{all the symptoms of depression}, so I am \textcolor{gray}{asking for help}}
      
      &
          \vspace{.25em}

      \textit{I think I am \textbf{clinically depressed} xxx. I have pretty much \textbf{every symptom on this list}.}   
          \vspace{1.5em}

      \\
      
      \midrule
      \vspace{.25em}
      \textit{... free of anxiety for a decade.. symptoms come ... }
      
      &
      
      \vspace{.25em}
      \textit{I've been \textbf{free of anxiety} \textcolor{gray}{for less than a week} and I'm afraid that it's only a temporary solution. \textcolor{gray}{I don't know what to do.}}
      
      &
      \vspace{.25em}
      \textit{I've been \textbf{free of anxiety for a decade} and I'm afraid that if the \textbf{symptoms start to come back}, I'll be able to \textbf{slip back into it} again.}
      
      \vspace{1.25em}

      \\
      \midrule
      
      
      \vspace{.25em}
      \textit{... ask her out ... barely talked to her... mess up my chance...}
      
      &
      \vspace{.25em}
      \textit{I like a girl \textcolor{black}{at work} and I want to \textbf{ask her out}, but I don't want to \textbf{ruin that chance} in the process.}
      
      &
      \vspace{.25em}
      \textit{I like a girl at work, but \textcolor{gray}{I don't know how to \textbf{ask her out.}}}
        \vspace{1.5em}
        \\      
      \bottomrule

    \end{tabular}
}
\caption{Four samples of the the human-written, \bart{}-generated, and \currsum-generated \tldr{} summaries. The human-written samples are partially shown to preserve users' privacy. That is, we have only shown the important human-written phrases to trace them within the generated summaries. The text that is unfaithful to the post (i.e., not supported by the user post) is in \textcolor{gray}{Gray} and the salient information that is picked up by the summarization systems is shown in \textbf{Bold}.}
\vspace{-1em}
\label{tab:er}
\end{table*}
\textbf{Automatic evaluation.}
Table \ref{tab:final} reports the performance of the baseline model along with our model's in terms of \textsc{Rouge} score variants~\cite{Lin2004ROUGEAP} and \textsc{Bertscore}~\cite{Zhang2020BertScore} over \mentsum dataset. As indicated, the best model is our \currsum{} that uses \superl{} curriculum directly on top of the \bart{} model and is a clear improvement over it across all metrics, achieving the current state-of-the-art performance. 
Specifically, our model outperforms its ground baseline that has no curriculum (i.e., \bart{}) by  improvement gains of 3.5\%, 10.4\%, 4.7\%, 1.5\% for \rgo, \rgt, \rgl, \textsc{bertscore},  respectively. Having looked at the \textsc{OracleExt} scores which shows the upper bound performance of an ideal extractive summarizer, it seems that there is room for improvement on the extractive setting to achieve state-of-the-art performance. More sophisticated models can invest in extractive or hybrid summarization models such as those done in ~\cite{Gehrmann2018BottomUpAS, MacAvaney2019SIG,Sotudeh2020AttendTM}.

\noindent \textbf{Case study and analysis.}
While our proposed model significantly improves upon the \bart{} baseline, we recognize the limitations of \textsc{rouge} metric in evaluation of summarization systems~\cite{Cohan2016RevisitingSE}. In order to explore the qualities and limitations of our work, we analyze the human-written \tldr{}s along with the generated results by \bart and our model, comparing them against each other. Table \ref{tab:er} shows four samples of the human and systems generated \tldr{}s. As seen, our model can improve the faithfulness of the summary~\footnote{Faithfulness is defined as generating output text that is supported by the user post.} in the first, second, and third samples. Having looked at other cases in our study, it appears that curriculum learning positively mitigates faithfulness errors. This might be attributed to the fact that the summarizer can achieve an improved \textit{understanding} of the source document when the contribution of each sample is controlled in each iteration of the learning process. Looking at the second sample, it turns out that our model can improve the conciseness of the summaries; that is, briefly conveying the main points within the summary. Comparing system-generated summaries in the fourth sample, it is observed that our system generated a phrase (shown in Gray) by mixing up different regions of the user post. Surprisingly, it appears that ``\textit{I don't know how/what to}'' is a common phrase used in most human-written summaries that are seeking advice from the community. The experimented summarization systems (i.e., \bart{} and ours) adhere to overgenerating this phrase. 

\section{Conclusion}
Generating short summaries given the users' online posts can save counselors' reading time, and reduce their fatigue. On this basis, they can provide faster responses to community users. While neural Transformers-based summarization models have shown to be promising, they suffer from \textit{inefficient training process} that hinders their potentials for showing a promising performance. To compensate for this issue, in this paper, 
we incorporated a confidence-aware curriculum learning  approach, which uses task-agnostic \textsc{SuperLoss}, to the summarization framework in the hope of increasing model's generalization, and ultimately improving model performance. Our automatic evaluations over \mentsum dataset of mental health posts show the effectiveness of our model, yielding 3.5\%, 10.4\%, 4.7\%, 1.5\% relative improvements over \bart{} summarizer on \rgo, \rgt, \rgl, and \textsc{bertscore}, respectively. Our model tailors the new state-of-the-art performance on \mentsum{} dataset. We further showed various system-generated summaries to showcase the qualities and limitations of our proposed model. 

\bibliography{main}
\bibliographystyle{acl_natbib}

\appendix



\end{document}